\documentclass[10pt,letterpaper,twocolumn]{article}

\usepackage{iccv}
\usepackage{times}

\usepackage{color,xcolor}
\usepackage{epsfig}
\usepackage{graphicx}

\usepackage{adjustbox}
\usepackage{array}
\usepackage{booktabs}
\usepackage{colortbl}
\usepackage{float,wrapfig}
\usepackage{hhline}
\usepackage{multirow}
\usepackage{subcaption} 
\usepackage[toc,page]{appendix}

\usepackage{amsmath,amsfonts,amsthm,amssymb}
\usepackage{bm}
\usepackage{nicefrac}
\usepackage{microtype}
\usepackage{inconsolata}
\usepackage{pifont}

\usepackage{changepage}
\usepackage{extramarks}
\usepackage{fancyhdr}
\usepackage{lastpage}
\usepackage{setspace}
\usepackage{soul}
\usepackage{xspace}
\usepackage{indentfirst}
\usepackage{paralist}

\usepackage[pagebackref=false,breaklinks=true,letterpaper=true,colorlinks,citecolor=citecolor,bookmarks=false]{hyperref}
\usepackage{url}

\usepackage{algorithm, algorithmic}
\usepackage{enumerate}
\usepackage{lipsum}
\newcolumntype{L}[1]{>{\raggedright\let\newline\\\arraybackslash\hspace{0pt}}m{#1}}
\newcolumntype{C}[1]{>{\centering\let\newline\\\arraybackslash\hspace{0pt}}m{#1}}
\newcolumntype{R}[1]{>{\raggedleft\let\newline\\\arraybackslash\hspace{0pt}}m{#1}}

\newcommand{\fig}[1]{Figure~\ref{#1}}

\newcommand{\tab}[1]{Table~\ref{#1}}

\newcommand{\ignorethis}[1]{}
\newcommand{\norm}[1]{\lVert#1\rVert}

\makeatletter
\DeclareRobustCommand\onedot{\futurelet\@let@token\@onedot}
\def\@onedot{\ifx\@let@token.\else.\null\fi\xspace}

\def\eg{\emph{e.g}\onedot} 
\def\ie{\emph{i.e}\onedot}

\def\etal{\emph{et al}\onedot}
\makeatother

\newcommand{\cmark}{\ding{51}}
\newcommand{\xmark}{\ding{55}}

\definecolor{citecolor}{HTML}{0071bc}
\definecolor{mydarkblue}{rgb}{0,0.08,1}
\definecolor{mydarkgreen}{rgb}{0.02,0.6,0.02}
\definecolor{mydarkred}{rgb}{0.8,0.02,0.02}
\definecolor{mydarkorange}{rgb}{0.40,0.2,0.02}
\definecolor{mypurple}{RGB}{111,0,255}
\definecolor{myred}{rgb}{1.0,0.0,0.0}
\definecolor{mygold}{rgb}{0.75,0.6,0.12}
\definecolor{mydarkgray}{rgb}{0.66, 0.66, 0.66}

\newcommand{\myparagraph}[1]{\vspace{-12pt}\paragraph{#1}}

\def\methodshort{LocTex\xspace}

\def\iccvPaperID{7238}

\iccvfinalcopy

\ificcvfinal\fi

\begin{document}

\title{\methodshort: Learning Data-Efficient Visual Representations \\ from \underline{Loc}alized \underline{Tex}tual Supervision}

\author{Zhijian Liu \\ MIT \and Simon Stent, Jie Li, John Gideon \\ Toyota Research Institute \\ \url{https://loctex.mit.edu/} \and Song Han \\ MIT}

\maketitle

\begin{abstract}

Computer vision tasks such as object detection and semantic/instance segmentation rely on the painstaking annotation of large training datasets. In this paper, we propose \methodshort that takes advantage of the low-cost \textbf{localized textual annotations} (\ie, captions and synchronized mouse-over gestures) to reduce the annotation effort. We introduce a contrastive pre-training framework between images and captions, and propose to supervise the cross-modal attention map with rendered mouse traces to provide coarse localization signals. Our learned visual features capture rich semantics (from free-form captions) and accurate localization (from mouse traces), which are very effective when transferred to various downstream vision tasks. Compared with ImageNet supervised pre-training, \methodshort can reduce the size of the pre-training dataset by \textbf{10$\times$} or the target dataset by \textbf{2$\times$} while achieving comparable or even improved performance on COCO instance segmentation. When provided with the same amount of annotations, \methodshort achieves around \textbf{4\%} higher accuracy than the previous state-of-the-art ``vision+language'' pre-training approach on the task of PASCAL VOC image classification.

\end{abstract}
\section{Introduction}

The tremendous success of deep learning in computer vision can be credited in part to the existence of large annotated datasets, such as ImageNet~\cite{deng2009imagenet,russakovsky2015imagenet}. However, acquiring high-quality annotations is usually very expensive and time-consuming, especially for dense, pixel-wise labeling tasks. For instance, segmenting instances in a single image from the COCO dataset takes more than 10 minutes on average~\cite{lin2014microsoft}\footnote{70k hours / 320k images = 0.22 hours/image = 13 minutes/image}.

Pre-training plus fine-tuning is a widely-adopted solution to reduce the need for costly annotations. In the computer vision community, a convolutional neural network (CNN) backbone is first pre-trained to perform image classification on ImageNet. Then, the learned features can be transferred to other downstream tasks by fine-tuning on the target dataset. Over the past few years, this paradigm has enabled state-of-the-art performance on many computer vision tasks, including object detection~\cite{ren2015faster}, semantic segmentation~\cite{long2015fully} and instance segmentation~\cite{he2017mask}.

\begin{figure}
\centering
\includegraphics[width=\linewidth]{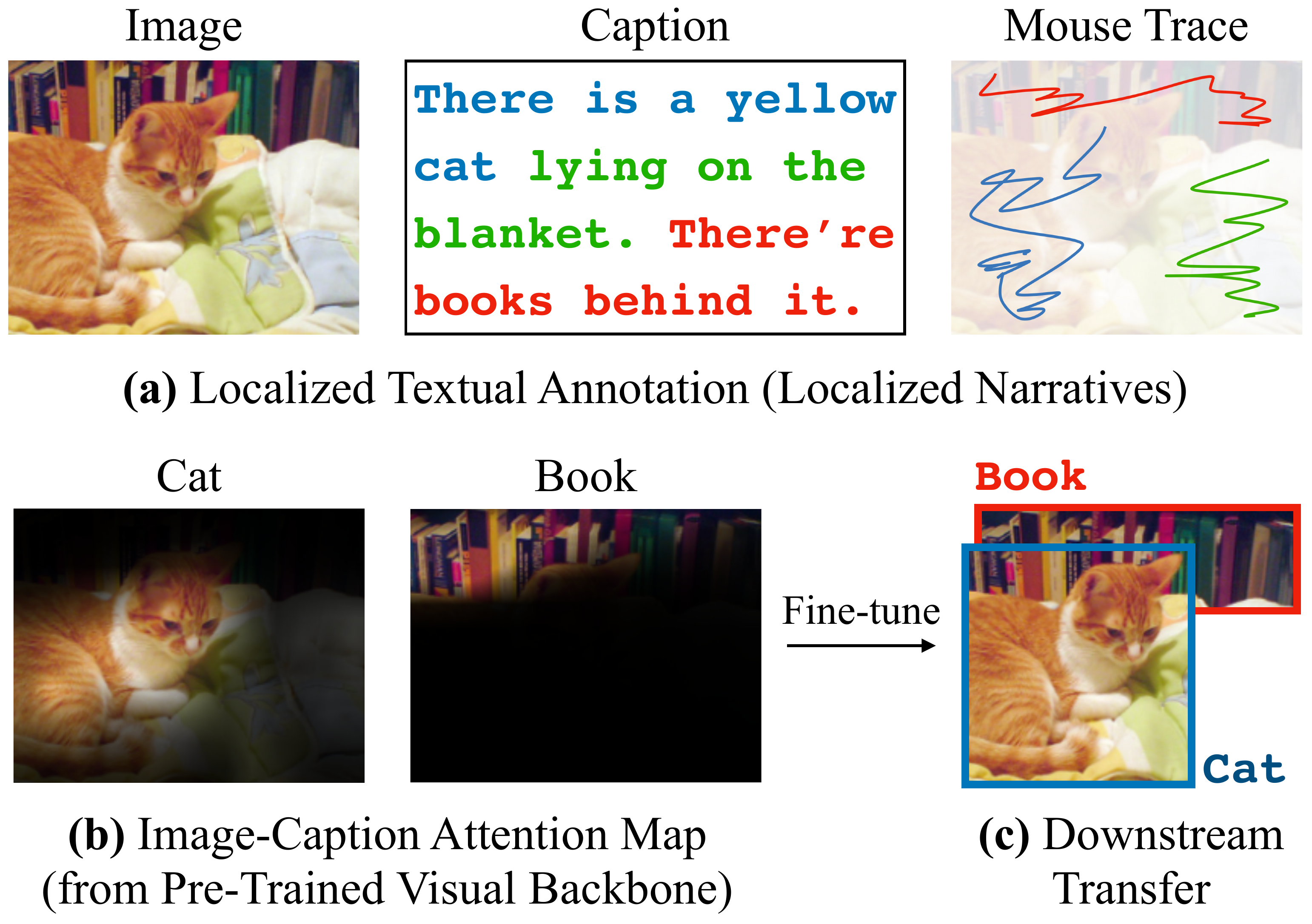}
\caption{\methodshort pre-trains the visual CNN backbone with (a) \emph{localized textual annotations}, which consists of free-form captions associated with synchronized mouse traces. With our contrastive and localization loss, the model learns (b) \emph{rich semantics and accurate localization}. This is very useful when transferred to (c) \emph{downstream tasks} that are sensitive to localization (\eg, object detection, instance segmentation).}
\label{fig:teaser}
\vspace{-12pt}
\end{figure}

Though effective, ImageNet pre-training has its caveats. \textbf{(i)} Its annotations (\ie, 1000-class labels) are very expensive to acquire. Annotating ImageNet is not as easy as it seems because differentiating among a fine-grained class taxonomy requires expert knowledge, which makes it hard to scale up or repeat. \textbf{(ii)} It is not as effective for those tasks that are more sensitive to localization than classification. As ImageNet pre-training only takes the object existence into consideration, its learned visual representations are supposed to be invariant to different object locations. Some recent research~\cite{he2019rethinking} has demonstrated competitive performance on object detection and instance segmentation with models trained from scratch. 

To solve (i), researchers have explored pre-training backbone networks with coarse, freely available labels, such as metadata and hashtags~\cite{joulin2016learning}. There has also been increased attention in self-supervised pre-training that learns visual representations from unlabeled images~\cite{he2020momentum,chen2020simple,grill2020bootstrap}. Some of them have been successfully scaled up to hundreds of millions or even billions of images~\cite{he2020momentum}. However, (ii) remains unsolved as they usually rely on some low-level visual cues (\eg, color, texture) and lack semantic understanding. In addition to this, \textbf{(iii)} self-supervised pre-training methods tend to be trained with prohibitively long schedules to exploit their potential. For instance, the recent approach of BYOL~\cite{grill2020bootstrap} requires 170 TPU days for a single training run.

In this paper, we propose \methodshort to learn data-efficient visual representations using \emph{localized textual supervision}, which is composed of free-form captions associated to synchronized mouse traces (see \fig{fig:teaser}a). This form of annotation can be easily acquired from non-expert workers, leading to (i) lower cost and better scalability. Technically, we propose to bridge the vision and language modalities with contrastive learning and supervise the cross-modal attention map with rendered mouse traces, providing (ii) coarse localization information that improves the performance of localization-sensitive downstream tasks. Finally, our method requires (iii) a similar amount of training time as ImageNet pre-training: it can be trained under a day with 8 GPUs.

After the pre-training, we transfer our learned feature representations to various downstream vision tasks, including image classification, object detection and instance segmentation. Compared with the ImageNet supervised pre-training, our proposed \methodshort can reduce the size of the pre-training dataset by \textbf{10$\times$} or the target dataset by \textbf{2$\times$} while achieving comparable or better performance on the COCO instance segmentation. With the same amount of annotations, our \methodshort achieves around \textbf{4\%} higher accuracy than the previous state-of-the-art ``vision+language'' pre-training approach~\cite{desai2021virtex} on the PASCAL VOC image classification.
\section{Related Work}

\paragraph{Supervised Pre-Training.}

Much of the recent success of computer vision can be attributed to the richness of image features learned via supervised training. ImageNet pre-training, in which image features are first learned through the supervised image classification on ImageNet~\cite{deng2009imagenet} before being used on downstream tasks, is a highly popular model initialization method~\cite{girshick2014rich, donahue2014decaf}. However, this approach has limitations which have become increasingly evident as the variety of downstream tasks and the types of new annotated data has increased dramatically over the years~\cite{raghu2019transfusion,he2019rethinking, zoph2020rethinking,shen2019object}.

\myparagraph{Unsupervised Learning.}

To go beyond the scale of ImageNet in terms of supervised learning is expensive. Hence, it becomes increasingly popular to seek methods for representation learning that can meet or exceed ImageNet supervised pre-training without the need for labelled data.
An important direction in unsupervised learning is ``self-supervised'' learning, in which models are trained on pretext tasks where training labels can be obtained from the raw or augmented input. Common pretext tasks include predicting context~\cite{doersch2015unsupervised}, solving jigsaws~\cite{noroozi2016unsupervised}, predicting rotation~\cite{gidaris2018unsupervised}, colorization~\cite{zhang2016colorful}, and inpainting~\cite{pathak2016context}. Generative models have also been widely used in representation learning to reconstruct the distribution of the input data, such as restricted Boltzmann machines (RBMs)~\cite{lee2009convolutional}, autoencoders~\cite{le2013building} and generative adversarial networks (GANs)~\cite{dumoulin2016adversarially,donahue2019large}. Recent explorations investigate intra-dataset patterns and feature discrimination, including clustering~\cite{caron2018deep,caron2020unsupervised} and contrastive learning~\cite{he2020momentum,grill2020bootstrap,chen2020simple,tian2019contrastive}.

\myparagraph{Vision \& Language.}

Pre-training methods in natural language processing have witnessed tremendous improvement over the past few years~\cite{NIPS2015_7137debd,peters2018deep,raffel2019exploring,brown2020language}. Efforts trying to use the text in visual representation learning have never stopped.
Early research tried to predict captions or text from associated images~\cite{quattoni2007learning}. Srivastava~\etal~\cite{srivastava2012multimodal} applied Boltzmann machine to capture multi-modal features.
Some works treat text or language as weak supervisory signals for vision and explore the trade-off between label quality and data scale. 
Li~\etal~\cite{li2017learning} train visual models on YFCC-100M~\cite{thomee2016yfcc100m} using user-provided tags. JFT-300M~\cite{sun2017revisiting} is also used for visual pre-training with automatic-generated web signals.
More recent works like ICMLM~\cite{sariyildiz2020learning}, VirTex~\cite{desai2021virtex} and ConVIRT~\cite{zhang2020contrastive} try to leverage the novel pre-training approaches developed in NLP, such as masked language modeling and transformer-based modeling. A concurrent work of us~\cite{radford2021learning} has explored contrastive learning between image and text at the web scale.
In this work, we explore further in this direction with a focus on learning localization-aware features for spatially-sensitive tasks such as object detection and segmentation.

\myparagraph{Annotation Efficiency.}

A key goal of our work is to learn powerful representations on data which can be acquired at a low annotation cost. Recent explorations focused on efficient labeling~\cite{acuna2018efficient,ling2019fast} and active learning~\cite{sener2018active} approach this problem by placing the model in the loop in order to increase the information gain per unit of annotation effort. We approach from an alternative angle by looking at widely available, natural sources of human supervision, which are already cheap or free to acquire. Our work centers around the Localized Narratives dataset~\cite{pont2020connecting}, which complements verbal image descriptions with synchronized mouse-over gestures containing noisy spatial cues. We demonstrate that designing a system to leverage such multi-modal cues can provide significant performance benefit on visual representation learning with minimal annotation overhead.
\begin{figure*}
\centering
\includegraphics[width=\linewidth]{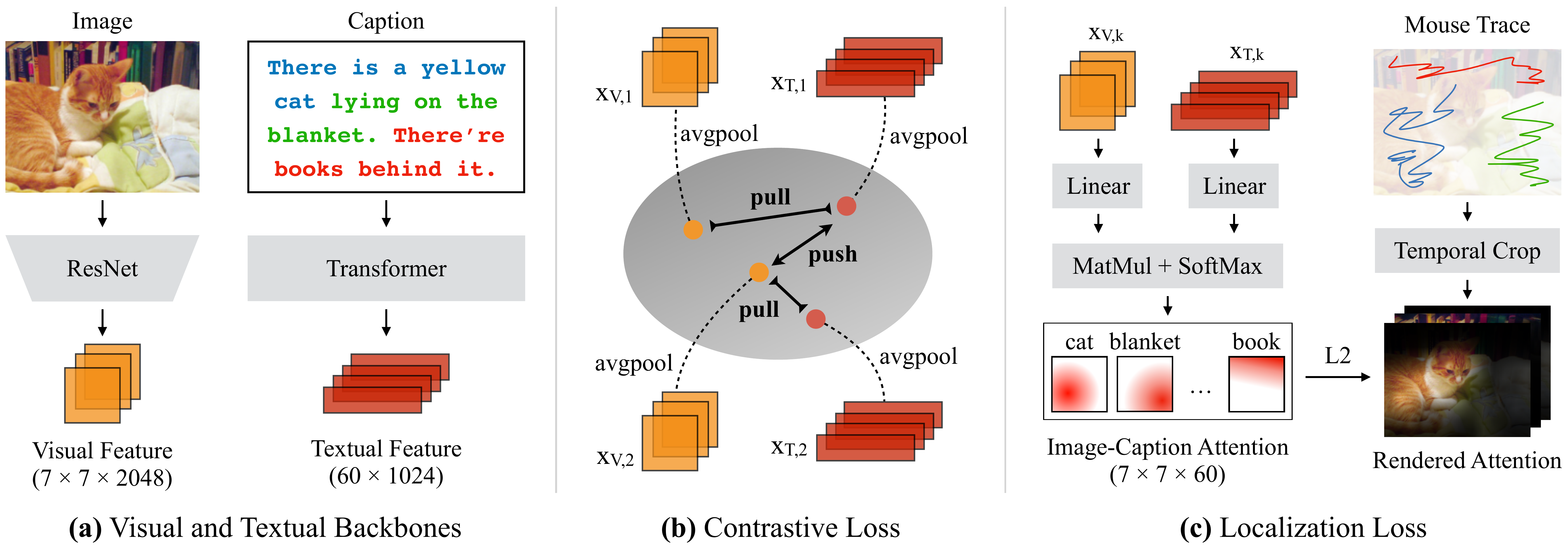}
\caption{Overview of our data-efficient visual representation learning framework (\methodshort). We first use a pair of visual and textual backbones to extract the features from the image and caption. We then apply the contrastive loss to pull the features from positive pairs together and push those from negative pairs apart. Finally, we compute the cross-modal attention map between visual and textual features and provide supervision using the rendered attention from the associated mouse trace.}
\label{fig:overview}
\vspace{-12pt}
\end{figure*}

\section{Method}
\label{sect:method}

In this section, we introduce our approach of visual representation learning from \emph{localized textual supervision}. We present an overview of our \methodshort framework in \fig{fig:overview}. We pre-train the visual backbone (as well as the textual backbone) using contrastive learning over positive and negative image-caption pairs. We propose to make use of the accompanying mouse trace annotations to provide coarse learning signals for localization. After pre-training, we transfer the learned visual backbone to other downstream vision tasks (\eg, classification, detection and segmentation).

\subsection{Annotations}

In the computer vision community, ImageNet~\cite{deng2009imagenet} was commonly used to pre-train visual backbone networks. However, annotating over 1000 fine-grained classes is very costly and cannot be easily scaled up~\cite{sariyildiz2020learning}. In this paper, we propose to employ \textit{localized textual annotations} (also known as localized narratives~\cite{pont2020connecting}) as it is relatively cheap to acquire and offers semantically dense information. The annotation we use consists of a caption with synchronized mouse trace:

\myparagraph{Caption.}

Caption is a free-form annotation resulting from annotators being asked to describe the content of the image using natural language. As illustrated in \fig{fig:teaser}a, the information captured in the caption is semantically dense: \ie, the objects in the image as well as their attributes and relative spatial relationships. The underlying rich semantic information could potentially benefit a variety of downstream vision tasks. On the other hand, the cost of this form of annotation is much lower compared with other dense labeling~\cite{lin2014microsoft} since it is a very natural task for humans to do and does not require the annotator to have extensive training or domain knowledge. Some recent datasets~\cite{pont2020connecting} adopt a two-stage data collection pipeline: they first ask the annotators to describe the image verbally and then apply either speech recognition or manual transcription to generate the final caption. From this collection protocol, the starting and ending timestamp of each token can also be obtained (which will be used to synchronize the mouse trace with the caption).

\myparagraph{Synchronized Mouse Trace.}

Compared with drawing a sequence of bounding boxes or instance masks, logging the mouse gestures of the subject while they describe the image is an easier and more natural way for human annotators to specify the object locations. It can be acquired almost freely in the caption annotation pipeline since the annotators only need to additionally hover their mouse over the region being described. Though the localization and semantic correspondence is too coarse for these annotations to be directly used for tasks like object detection, it does capture rich information about ``what is where'' at a high level.

\subsection{Backbones}

Given an image and its corresponding caption, we first apply two separate neural networks to extract their features.

\myparagraph{Visual Backbone.}

The visual backbone takes the raw image as input and outputs a feature map that contains the semantic information. This is also the only component which we will transfer to other vision downstream tasks. Theoretically, we can choose any convolutional neural network as our visual backbone. Following recent representation learning papers~\cite{he2020momentum,chen2020simple,desai2021virtex}, we adopt a standard ResNet-50~\cite{he2016deep} as our feature extractor throughout this paper to facilitate fair comparison. We remove the last linear classification layer and the preceding global average pooling layer to keep the spatial dimension. Thus, the output feature map from the visual backbone will have size $\text{2048} \times R \times R$, where $R$ is the output resolution (which is 1/32 of the input resolution).

\myparagraph{Textual Backbone.}

The textual backbone encodes the input caption into a feature vector that captures the meaning of each word token. In this paper, we adopt a Transformer~\cite{vaswani2017attention} architecture as our textual backbone. Specially, we implement a 4-layer 1024-wide model with 16 self-attention heads. Similar to Desai~\etal~\cite{desai2021virtex}, we replace the activation function from ReLU to GELU~\cite{hendrycks2016gaussian} for its better empirical performance. We refer the readers to Vaswani~\etal~\cite{vaswani2017attention} for more architectural details. Before feeding the caption in, we first tokenize it into a lower-cased byte pair encoding (BPE)~\cite{sennrich2016neural} with a vocabulary size of 10K. This results in almost no out-of-vocab unknown (\texttt{[UNK]}) tokens in our experiments. We also pad the input sequence with start of sequence (\texttt{[SOS]}) and end of sequence (\texttt{[EOS]}) tokens to mark the boundary. The output feature vector from the textual backbone has size $\text{1024} \times L$ where $L$ is the caption length after tokenization.

\subsection{Contrastive Loss}

Given a batch of feature pairs extracted from visual and textual backbones: $\{(\bm{x}_{\text{V},k}, \bm{x}_{\text{T},k}) \mid 1 \leq k \leq n\}$ (where $n$ is the batch size), we transform each feature with a global average pooling and a single 1024-dimension fully-connected layer. The resulting visual and textual features are denoted $\bm{y}_{\text{V},k}$ and $\bm{y}_{\text{T},k}$ (both size 1024). Now, a straightforward way to guide the pre-training is to match $\bm{y}_{\text{V},k}$ and $\bm{y}_{\text{T},k}$ in the feature space using a simple L1/L2 regression loss. However, this will lead to a collapsed solution where all features are projected to the same location in the feature space~\cite{grill2020bootstrap}.

Motivated by Chen~\etal~\cite{chen2020simple}, we encourage the visual and textual backbones to not only project the features of \emph{matching} image-caption pairs to be \emph{closer} but also the features of \emph{non-matching} pairs to be \emph{further}.  Concretely, there are $n^2$ image-caption pairs $\{(\bm{y}_{\text{V},i}, \bm{y}_{\text{T},j}) \mid 1 \leq i, j \leq n\}$ in total, among which only the $n$ pairs with $i = j$ are positive (as they correspond to the same data) while the remaining $(n^2-n)$ pairs are negative. We use the InfoNCE loss~\cite{oord2018representation} to pull the positive pairs together and push the negative pairs apart to guide the pre-training (see \fig{fig:overview}b):
\begin{equation}
\label{eqn:loss:c}
    \mathcal{L}_\text{C} = -\sum_{i=1}^n \log \frac{\exp(\text{sim}(\bm{y}_{\text{V},i}, \bm{y}_{\text{T},i}) / \tau)}{\sum_{j \neq i}{\exp(\text{sim}(\bm{y}_{\text{V},i}, \bm{y}_{\text{T},j}) / \tau)}},
\end{equation}
where $\text{sim}(\bm{u}, \bm{v}) = \bm{u}^\text{T}\bm{v} / \norm{\bm{u}}_2\norm{\bm{v}}_2$ is the cosine similarity between two vectors, and $\tau$ denotes a temperature parameter (which is set to 0.1 in our experiments).

\myparagraph{Discussions.}

Contrastive learning is not the only way to bridge the vision and language modalities. It is also possible to use one modality as input and the other as output to form a supervised learning problem: \ie, image captioning~\cite{desai2021virtex} (image to caption) and image synthesis~\cite{ramesh2021zero} (caption to image). However, the supervised formulation has a worse empirical performance than the contrastive one (see our comparisons with VirTex~\cite{desai2021virtex} in \tab{tab:analysis:ablations}). Similar observations have also been made in our concurrent work~\cite{radford2021learning}. We conjecture that this is because the relationship between image and caption is not one-to-one (\ie, a single image may be described in a multitude of ways, and vice versa). In this case, the encoding process (many-to-one projection) might be much easier than the decoding process (one-to-many projection).

\subsection{Localization Loss}

Applying the contrastive loss over the global visual and textual features (after average pooling) provides the model with a holistic sense of what objects are in the image. However, the model might not be able to correspond each instance with its spatial location. This greatly limits its effectiveness when transferred to localization-sensitive downstream tasks (\eg, object detection, instance segmentation). This is where the mouse trace can be helpful since it provides coarse localization information about the instances: \ie, where the annotators position their mouse when describing an object.

We provide an overview of our localization loss in \fig{fig:overview}c. We first transform visual and textual features linearly using a 1024-dimension fully-connected layer. Note that we do not apply the global average pooling as we need to keep the spatial dimension to learn localization. Thus, the transformed visual feature $\bm{z}_{\text{V},k}$ will have a size of $\text{1024} \times R \times R$, and the transformed textual feature $\bm{z}_{\text{V},k}$ will have a size of $\text{1024} \times L$. We then compute the image-caption attention map as the normalized product between two feature maps:
\begin{equation}
    \bm{\mathcal{M}}_k = \text{softmax}(\bm{z}_\text{T, k}^\text{T} \times \bm{z}_\text{V, k}),
\end{equation}
which will then have the size of $L \times R \times R$. In $\bm{\mathcal{M}}_k$, each location $(i, x, y)$ corresponds to (the probability of) whether the object described by the token $i$ is located in the region of $(x, y)$. We observe that this may be supervised using the mouse trace.

Given the fact that the mouse trace is synchronized with the caption, we first temporally crop the part of the mouse trace sequence that corresponds to each token in the caption. We then render the covered region of cropped mouse trace into a binary mask with resolution of $R$. Finally, we stack the rendered masks of all tokens together to generate the rendered attention $\hat{\bm{\mathcal{M}}}_k$. Since it has the same format and definition as the image-caption attention map $\bm{\mathcal{M}}_k$, we can use it to provide supervision on $\bm{\mathcal{M}}_k$ with a normalized L2 regression loss:
\begin{equation}
    \mathcal{L}_\text{L} = \sum_{k=1}^n \left\lVert \bm{\mathcal{M}}_k / \norm{\bm{\mathcal{M}}_k}_2 - \hat{\bm{\mathcal{M}}}_k / \norm{\hat{\bm{\mathcal{M}}}_k}_2 \right\rVert_2.
\end{equation}

\myparagraph{Discussions.}

The feature map from the visual backbone usually has a low resolution (\ie, $R=7$ if the input size is 224$\times$224), which largely limits the precision of the provided localized supervision. Therefore, we additionally apply the localization loss to the second last visual feature maps (which has 2$\times$ larger resolution) to provide supervision at a finer scale. The losses computed at different resolutions are added together with equal weights. We note that using even higher resolutions than this leads to worse performance (see \tab{tab:analysis:supervisions}). A likely reason for this is that the mouse trace annotations from the datasets we use, and mouse traces in general, are intrinsically noisy. In this case, downsampling to a lower resolution removes some of the spurious correlations that otherwise might be introduced, at the cost of weaker overall supervision.

\subsection{Implementation Details}

\paragraph{Pre-Training Dataset.}

We use Localized Narratives~\cite{pont2020connecting} as our pre-training dataset as it provides large-scale localized textual annotations: \ie, it annotates the whole COCO~\cite{lin2014microsoft}, Flickr30k~\cite{young2014image}, ADE20k~\cite{zhou2019semantic}, and part of Open Images~\cite{kuznetsova2020open} datasets with high-quality captions and synchronized mouse traces. In this paper, we present two variants of our \methodshort: (1) a smaller one trained only with COCO images (which contains 118K images) to have a fair comparison with other ``vision+language'' baselines, and (2) a larger one trained on both COCO and Open Images data (which contains 809K annotated images) to test the scalability of our method. To compensate for the resolution difference with COCO, we downsample the images from Open Images by 0.6$\times$.

\myparagraph{Data Augmentation.}

We apply standard data augmentations for images: \ie, random crop, random horizontal flip, color jittering and normalization. Following Desai~\etal~\cite{desai2021virtex}, we swap the `left' and `right' tokens in the caption when applying the horizontal flip. We limit the caption length to 60 tokens for computational efficiency: we pad the caption with zeros if its length is shorter than 60 or otherwise crop a random 60-token subsequence from the caption, which empirically helps to reduce overfitting.

\myparagraph{Loss Functions.}

We assign $\mathcal{L}_\text{C}$ and $\mathcal{L}_\text{L}$ with equal weights as they are roughly of the same magnitude. The contrastive loss is computed locally at each GPU to save the communication bandwidth. This reduces the number of negative pairs, while empirically, the convergence rate is not affected.

\myparagraph{Training Details.}

We pre-train the visual and textual backbones with a batch size of 1024 for 600 epochs. Optimization is carried out using stochastic gradient descent with a momentum of 0.9 and a weight decay of $10^{-4}$. We use a learning rate of 0.4 for the visual backbone, 0.002 for the textual backbone, and 0.4 for the linear transforms. We adopt the cosine learning rate decay schedule~\cite{loshchilov2017sgdr} with a linear warmup for the first 20 epochs. We distribute the training over 8 NVIDIA V100 GPUs with synchronized batch normalization~\cite{peng2018megdet} and automatic mixed-precision~\cite{micikevicius2018mixed} (from PyTorch~\cite{paszke2019pytorch}). The total training time is around 18 hours.
\section{Experiments}

\begin{table}[!t]
\setlength{\tabcolsep}{7pt}
\small\centering
\begin{tabular}{lC{38pt}cc}
    \toprule
     & \# Pretrain Images & Annotations & mAP \\
    \midrule
    Random Init & -- & -- & 67.3 \\
    \midrule
    MoCo~\cite{he2020momentum} & 1.28M & self-supervised & 79.4 \\
    PCL~\cite{li2021prototypical} & 1.28M & self-supervised & 83.1 \\
    SwAV~\cite{caron2020unsupervised} & 1.28M & self-supervised & 87.9 \\
    IN-Sup & 1.28M & 1 (1000-class) label & 86.8 \\
    \midrule
    VirTex~\cite{desai2021virtex} & 118K & 1 caption & 84.2 \\
    \methodshort (Ours) & 118K & 1 localized caption & \textbf{88.4} \\
    \midrule
    ICMLM~\cite{sariyildiz2020learning} & 118K & 5 captions & 87.5 \\
    VirTex~\cite{desai2021virtex} & 118K & 5 captions & 88.7 \\
    \methodshort (Ours) & 809K & 1 localized caption & \textbf{92.6} \\
    \bottomrule
\end{tabular}
\caption{Results of linear classification on PASCAL VOC. Our \methodshort outperforms supervised and self-supervised pre-training on ImageNet by \textbf{4-13\%} while using around 60\% of the annotated images. It also achieves \textbf{4\%} higher accuracy than previous vision+language pre-training methods when trained with a similar amount of annotations.}
\label{tab:results:voc_classification}
\vspace{-12pt}
\end{table}

In this section, we evaluate the effectiveness of our pre-trained visual backbone in various downstream vision tasks, including image classification, object detection and instance segmentation. The textual backbone also learns useful representations and can be transferred to language-related tasks in principle, though exploration of this is left as future work.

\subsection{Image Classification}

\begin{table*}[!t]
\setlength{\tabcolsep}{4pt}
\small\centering
\begin{tabular}{lC{38pt}cccccccccccc}
    \toprule
     & & \multicolumn{6}{c}{10\% Training Data} & \multicolumn{6}{c}{20\% Training Data} \\
    \cmidrule(lr){3-8}\cmidrule(lr){9-14}
     & \# Pretrain Images & $\text{AP}^\text{bbox}$ & $\text{AP}^\text{bbox}_\text{50}$ & $\text{AP}^\text{bbox}_\text{75}$ & $\text{AP}^\text{mask}$ & $\text{AP}^\text{mask}_\text{50}$ & $\text{AP}^\text{mask}_\text{75}$ & $\text{AP}^\text{bbox}$ & $\text{AP}^\text{bbox}_\text{50}$ & $\text{AP}^\text{bbox}_\text{75}$ & $\text{AP}^\text{mask}$ & $\text{AP}^\text{mask}_\text{50}$ & $\text{AP}^\text{mask}_\text{75}$ \\
    \midrule
    Random Init & -- & 16.0 & 29.6 & 15.3 & 15.1 & 27.3 & 15.0 & 17.8	& 31.7 & 17.8 & 16.7 & 29.6 & 17.0 \\
    \midrule
    IN-Sup (10\%) & 128K & 16.4 & 31.7 & 15.3 & 15.7 & 29.1 & 15.4 & 22.3 & 39.2 & 22.5 & 20.8 & 36.5 & 21.2 \\
    VirTex~\cite{desai2021virtex} & 118K & 23.7 & 41.9 & 24.0 & 21.5 & 38.6 & 21.4 & 28.9 & 47.4 & 30.6 & 25.6 & 44.1 & 26.2 \\
    \methodshort (Ours) & 118K & \textbf{25.0} & \textbf{43.2} & \textbf{25.7} & \textbf{22.4} & \textbf{39.8} & \textbf{22.4} & \textbf{29.8} & \textbf{48.9} & \textbf{31.1} & \textbf{26.4} & \textbf{45.2} & \textbf{27.2} \\
    \midrule
    IN-Sup (50\%) & 640K & 23.4 & 41.9 & 23.5 & 21.6 & 38.5 & 21.6 & 28.5 & 47.3 & 29.8 & 25.5 & 43.9 & 26.5 \\
    VirTex~\cite{desai2021virtex} & 118K($\times$5) & 26.3 & 44.1 & 27.1 & 23.4 & 40.9 & 23.8 & 30.7 & 49.4 & 32.3 & 27.1 & 45.9 & 27.9 \\
    \methodshort (Ours) & 809K & \textbf{27.3} & \textbf{45.8} & \textbf{28.2} & \textbf{24.2} & \textbf{42.1} & \textbf{24.9} & \textbf{31.8} & \textbf{50.9} & \textbf{33.8} & \textbf{27.8} & \textbf{47.3} & \textbf{28.9} \\
    \midrule
    IN-Sup (100\%) & 1.28M & 25.0 & 43.8 & 25.2 & 22.8 & 40.1 & 23.0 & 30.3	& 49.9 & 31.6 & 27.0 & 46.1 & 27.9 \\
    \bottomrule
\end{tabular}
\caption{Results of instance segmentation on COCO. Our \methodshort consistently outperforms VirTex and IN-Sup under 10\% and 20\% data settings. We refer the readers to the appendix for detailed results under 50\% and 100\% data settings.}
\label{tab:results:coco}
\vspace{-12pt}
\end{table*}

Following the common protocol~\cite{he2020momentum}, we first evaluate our method by linear classification on frozen features: the pre-trained visual backbone is fixed and used to extract features.

\myparagraph{Setup.}

We adopt the PASCAL VOC dataset~\cite{everingham2015pascal} for our linear evaluation. We first resize all images to 224$\times$224 and feed them into our pre-trained ResNet-50. We then apply global average pooling to extract 2048-dimensional image features. We train a separate SVM for each class on VOC07 \texttt{trainval} and report the mean AP (over 20 classes) on the \texttt{test} split. Following VirTex~\cite{desai2021virtex}, we train multiple SVMs with different cost values from $\{0.01, 0.1, 1, 10\}$ and select the best SVM based on a 3-fold cross-validation.

\myparagraph{Baselines.}

We compare our method with three sets of baselines: (1) ImageNet pre-training (IN-Sup) that pre-trains the model on the large-scale ImageNet dataset to perform image classification, (2) self-supervised learning~\cite{he2020momentum,li2021prototypical,caron2020unsupervised} that pre-trains the model with a large number of unlabeled images, and (3) vision+language pre-training~\cite{sariyildiz2020learning,desai2021virtex} that pre-trains the model to perform image captioning on COCO.

\myparagraph{Results.}

Training the classifier from scratch yields a rather poor performance because the size of PASCAL VOC is fairly small (with only 9K images). The widely-adopted ImageNet pre-training (IN-Sup) significantly boosts the accuracy; however, it requires massive annotations over a fine-grained class hierarchy. From \tab{tab:results:voc_classification}, our \methodshort achieves 1.6\% higher accuracy than IN-Sup with only \textbf{10\%} of annotated images, or \textbf{5.8\%} higher with around 60\% of annotated images. The superior performance comes from the use of cheap yet semantically dense localized caption annotations.

Previous vision+language pre-training methods~\cite{sariyildiz2020learning,desai2021virtex} were trained with five captions per image, which increases the annotation cost by 5$\times$. To have a fair comparison, we compare our \methodshort with the 1-caption VirTex~\cite{desai2021virtex}. With the same amount of pre-training images, our \methodshort achieves more than \textbf{4\%} higher accuracy, which is contributed by the better optimization formulation and the additional localization supervision (see \tab{tab:analysis:ablations}). We are also on par in terms of the annotation cost as the extra mouse trace annotations we use can be acquired almost for free during the caption annotation~\cite{pont2020connecting}. We further scale our method up with the additional Open Images data. With a similar amount of annotated images, our \methodshort outperforms the full VirTex by \textbf{4\%} and ICMLM~\cite{sariyildiz2020learning} by around \textbf{5\%}.

\subsection{Object Detection}

We then evaluate our method by transferring our learned visual backbone to object detection. Here, the entire backbone is fine-tuned along with the object detector.

\myparagraph{Setup.}

We adopt the PASCAL VOC dataset~\cite{everingham2015pascal} for our detection evaluation. Different from the linear evaluation setup, we also include VOC12 \texttt{trainval} into the training set. For the object detector, we use Faster-RCNN~\cite{ren2015faster} with ResNet-C4 backbone. Following He~\etal~\cite{he2020momentum}, we add an extra batch normalization right after the visual backbone. We fine-tune all models for 24K iterations with linear warmup. The learning rate is initialized with 0.02 and decayed by 10$\times$ at 18K and 22K iteration. We distribute the training across 8 GPUs with a total batch size of 16.

\begin{table}[!t]
\setlength{\tabcolsep}{7pt}
\small\centering
\begin{tabular}{lC{38pt}ccc}
    \toprule
     & \# Pretrain Images & $\text{AP}^\text{bbox}$ & $\text{AP}^\text{bbox}_\text{50}$ & $\text{AP}^\text{bbox}_\text{75}$ \\
    \midrule
    Random Init & -- & 33.8 & 60.2 & 33.1\\
    \midrule
    IN-Sup (10\%) & 128K & 42.6 & 72.0 & 43.8 \\
    MoCo~\cite{he2020momentum} & 118K & 47.6 & 75.4 & 51.0 \\
    VirTex~\cite{desai2021virtex} & 118K & 51.7 & 79.6 & 56.5\\
    \methodshort (Ours) & 118K & \textbf{53.9} & \textbf{80.9} & \textbf{59.8} \\
    \midrule
    IN-Sup (50\%) & 640K & 52.1 & 80.4 & 57.0 \\
    VirTex~\cite{desai2021virtex} & 118K($\times$5) & 55.3 & 81.3 & 61.0\\
    \methodshort (Ours) & 809K & \textbf{56.9} & \textbf{82.4} & \textbf{63.2} \\
    \midrule
    IN-Sup (100\%) & 1.28M & 54.3 & 81.4 & 59.6 \\
    \bottomrule
\end{tabular}
\caption{Results of object detection on PASCAL VOC. Our \methodshort surpasses VirTex and IN-Sup by \textbf{1.5-2.2\%} and \textbf{4.8-11.3\%} given a similar amount of pre-training images.}
\label{tab:results:voc_detection}
\vspace{-12pt}
\end{table}
\begin{figure*}[!t]
\centering
\includegraphics[width=\linewidth]{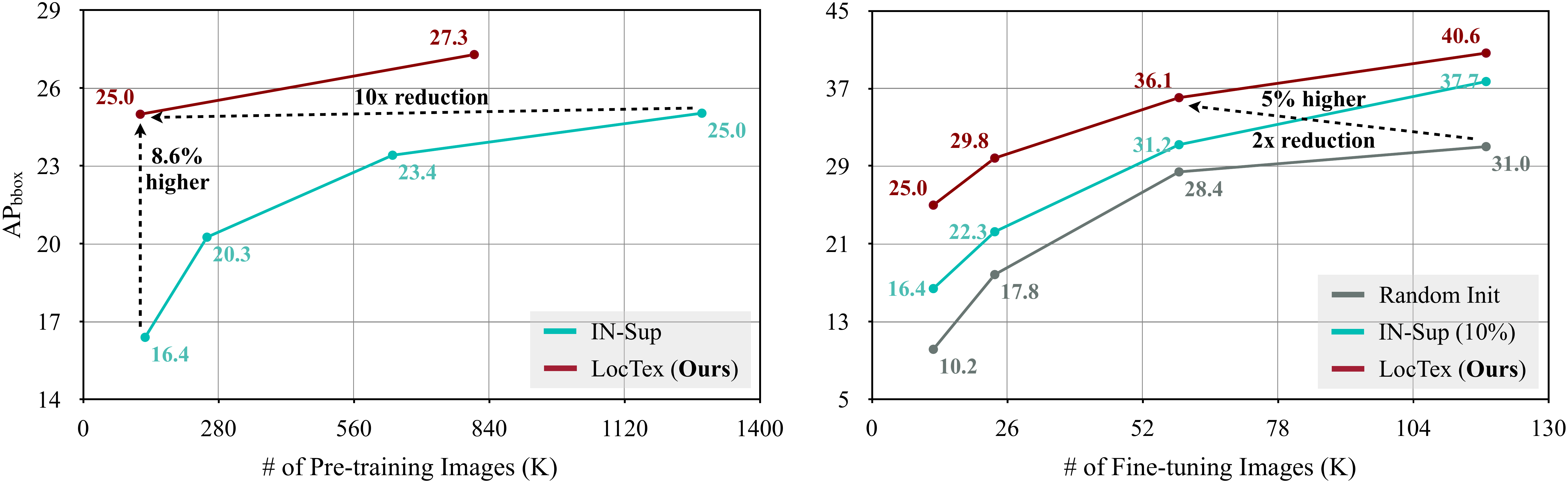}
\caption{\methodshort learns visual representations in a data-efficient manner: on COCO instance segmentation, it is able to reduce the pre-training dataset by \textbf{10$\times$} without loss of accuracy or reduce the target dataset by \textbf{2$\times$} with \textbf{5\%} higher accuracy.}
\label{fig:results:coco_tradeoffs}
\vspace{-12pt}
\end{figure*}

\myparagraph{Baselines.}

Apart from the full ImageNet pre-training baseline, we also scale it down with fewer pre-training images (10\%, 20\%, 50\%) to match the annotation cost of VirTex and ours. We follow the same training protocol as \texttt{torchvision} and keep the number of epochs the same; otherwise, these models trained on smaller subsets are more prone to overfitting. These baselines are referred to as IN-Sup ($k$\%).

\myparagraph{Results.}

We present our object detection results in \tab{tab:results:voc_detection}. With a similar amount of pre-training images, our \methodshort surpasses VirTex and IN-Sup by a large margin (\textbf{1.5-2.2\%} and \textbf{4.8-11.3\%}, respectively). Remarkably, \methodshort matches the full ImageNet pre-training performance with more than \textbf{10$\times$} fewer annotated images. The scaled-up version of \methodshort further pushes the AP to 56.9\%, which is 2.6\% higher than the full ImageNet pre-training performance despite using \textbf{1.6$\times$} fewer images.

\subsection{Instance Segmentation}

Finally, we evaluate our method on instance segmentation under the \emph{limited data} setting. Similar to the detection setup, we train the visual backbone end-to-end with the model.

\myparagraph{Setup.}

We use the COCO dataset~\cite{lin2014microsoft} (with \texttt{train2017} and \texttt{val2017} split) for segmentation evaluation. We choose Mask R-CNN~\cite{he2017mask} with ResNet-C4 backbone as our model. We add the extra batch normalization to the visual backbone. Following the 2$\times$ schedule, we train the model with 180K iterations. The learning rate is initialized with 0.02, multiplied by 0.01 at 120K and 160K iteration. As we target at the limited data setting, we sample a subset from COCO images (\eg, 10\%, 20\%, 50\%, 100\%) for fine-tuning, and shrink the training schedule proportionally to the dataset size.

\myparagraph{Results.}

From \tab{tab:results:coco}, our proposed \methodshort consistently outperforms VirTex and IN-Sup under all data settings. We refer the readers to the appendix for detailed results under 50\% and 100\% data settings. In \fig{fig:results:coco_tradeoffs}, we further investigate our method from the data efficiency perspective:
\begin{itemize}[\hspace{1pt}--]
  \item \textbf{Pre-Training Data.} Our \methodshort can reduce the number of pre-training images by \textbf{10$\times$} without loss of accuracy. With the same amount of pre-training data, it outperforms IN-Sup by more than \textbf{8\%} in terms of AP. This translates into \textbf{2.4$\times$} and \textbf{6.4$\times$} lower annotation cost compared to pre-training with classification and segmentation labels. We refer the readers to the appendix for more details.
  \item \textbf{Fine-Tuning Data.} The end goal of a good pre-training is to reduce the amount of costly annotation in the target task. Our \methodshort reduces the target dataset by \textbf{2$\times$} while achieving more than \textbf{5\%} higher accuracy than training from scratch. Under extremely limited data settings (\ie, 5-10\%), the improvement is even more significant: \textbf{2.7\%} and \textbf{7.2\%} AP boost compared with ImageNet pre-training and random initialization, with \textbf{2$\times$} data reduction.
\end{itemize}

\section{Analysis}

In this section, we provide some additional analysis of our model to understand how it works and might be improved.

\myparagraph{Effectiveness of $\mathcal{L}_\text{C}$ and $\mathcal{L}_\text{L}$.}

\begin{table}[t]
\setlength{\tabcolsep}{7pt}
\small\centering
\begin{tabular}{ccccccc}
    \toprule
    & & VOC & \multicolumn{2}{c}{COCO (10\%)} & \multicolumn{2}{c}{COCO (20\%)} \\
    \cmidrule(lr){3-3}\cmidrule(lr){4-5}\cmidrule(lr){6-7}
    $\mathcal{L}_\text{C}$ & $\mathcal{L}_\text{L}$ & mAP & $\text{AP}^\text{bbox}$ & $\text{AP}^\text{mask}$ & $\text{AP}^\text{bbox}$ & $\text{AP}^\text{mask}$ \\
    \midrule
    \cmark & \cmark & 88.4 & 25.0 & 22.4 & 29.8 & 26.4 \\
    \midrule
    \cmark & \xmark & \textcolor{darkgray}{--0.9} & \textcolor{darkgray}{--0.7} & \textcolor{darkgray}{--0.6} & \textcolor{darkgray}{--0.5} & \textcolor{darkgray}{--0.5} \\
    \xmark & \xmark & \textcolor{darkgray}{--4.2} & \textcolor{darkgray}{--1.3} & \textcolor{darkgray}{--0.9} & \textcolor{darkgray}{--0.9} & \textcolor{darkgray}{--0.8} \\
    \bottomrule
\end{tabular}
\caption{The formulation of contrastive learning ($\mathcal{L}_\text{C}$) and the use of low-cost mouse trace annotations ($\mathcal{L}_\text{L}$) are important to the effectiveness of our visual representation learning.}
\label{tab:analysis:ablations}
\vspace{-12pt}
\end{table}
\begin{figure*}
\centering
\includegraphics[width=0.136\textwidth]{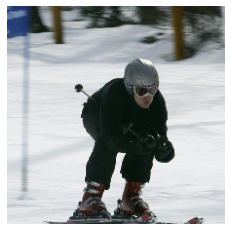}
\includegraphics[width=0.136\textwidth]{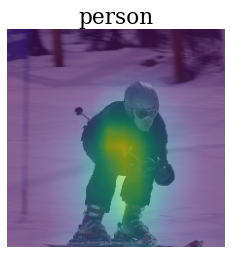}
\includegraphics[width=0.136\textwidth]{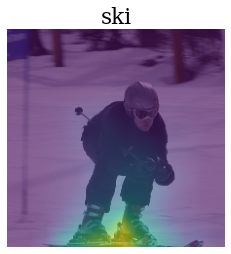}
\includegraphics[width=0.136\textwidth]{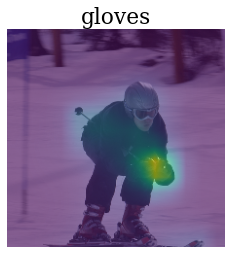}
\includegraphics[width=0.136\textwidth]{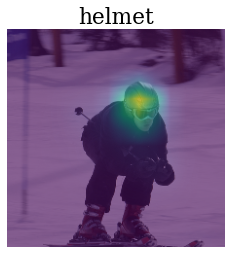}
\includegraphics[width=0.136\textwidth]{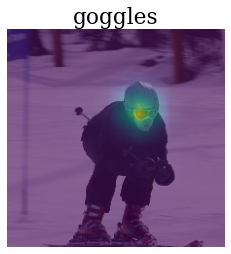}
\includegraphics[width=0.136\textwidth]{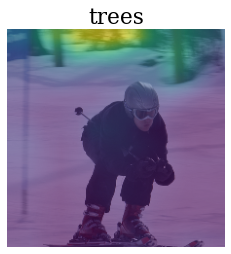}

\includegraphics[width=0.136\textwidth]{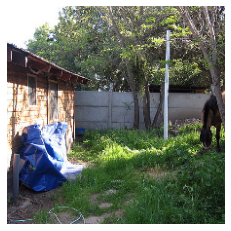}
\includegraphics[width=0.136\textwidth]{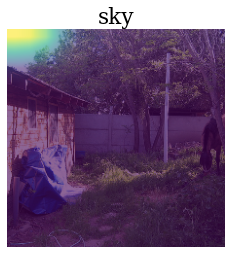}
\includegraphics[width=0.136\textwidth]{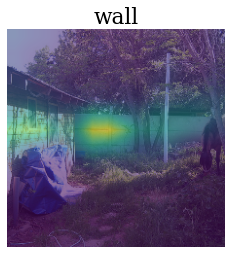}
\includegraphics[width=0.136\textwidth]{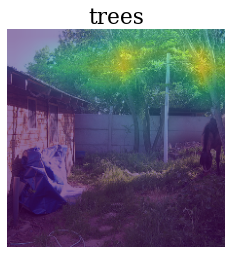}
\includegraphics[width=0.136\textwidth]{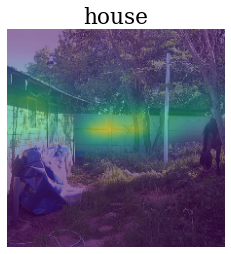}
\includegraphics[width=0.136\textwidth]{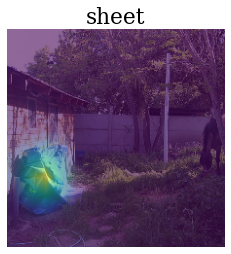}
\includegraphics[width=0.136\textwidth]{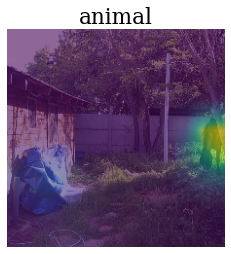}

\includegraphics[width=0.136\textwidth]{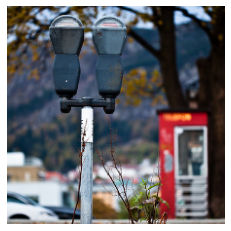}
\includegraphics[width=0.136\textwidth]{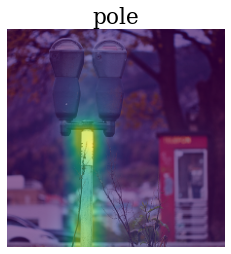}
\includegraphics[width=0.136\textwidth]{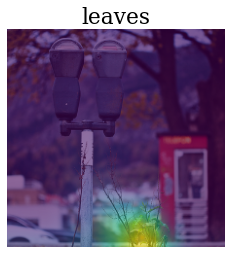}
\includegraphics[width=0.136\textwidth]{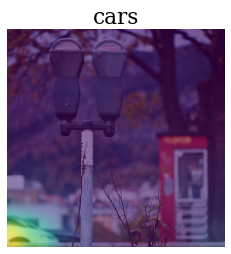}
\includegraphics[width=0.136\textwidth]{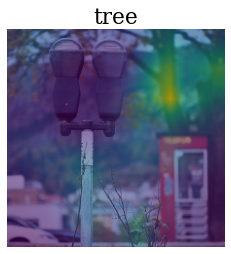}
\includegraphics[width=0.136\textwidth]{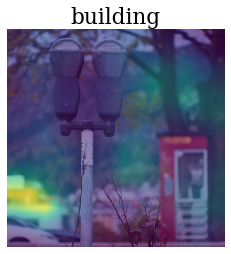}
\includegraphics[width=0.136\textwidth]{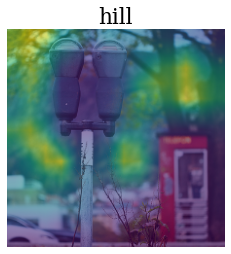}

\caption{Visualization of learned image-caption attention maps (on COCO \texttt{val2017}). Our \methodshort learns rich visual concepts (\eg, helmets, goggles) and fairly accurate localization. We refer the readers to the appendix for further examples.}
\label{fig:analysis:attentions}
\vspace{-12pt}
\end{figure*}

The two major components of \methodshort are the formulation of contrastive learning ($\mathcal{L}_\text{C}$) and the use of low-cost mouse trace annotations ($\mathcal{L}_\text{L}$). Thus, we present some ablation analysis by removing one or both from our framework. VirTex~\cite{desai2021virtex} can be seen as our model removing both $\mathcal{L}_\text{C}$ and $\mathcal{L}_\text{L}$ (and using a predictive loss instead). From \tab{tab:analysis:ablations}, both components contributes positively to our final performance on downstream vision tasks. We also observe that the contrastive loss is particularly effective on image classification; while the localization loss is more useful on instance segmentation. This phenomenon is well aligned with our design where $\mathcal{L}_\text{C}$ provides holistic semantic information and $\mathcal{L}_\text{L}$ offers detailed localization supervision.

\myparagraph{Learned Image-Caption Attention Map.}

Although we focus on transferring the learned visual backbone to different downstream tasks, it is still fairly important to understand what the model actually learns from the pre-training stage. In \fig{fig:analysis:attentions}, we visualize the learned image-caption attention map. We refer the readers to the appendix for more examples. Here, the visualized attention maps are predicted from the second last visual feature map (with resolution of 14$\times$14). We resize the attention maps to 224$\times$224 and then overlay them to images. As shown in \fig{fig:analysis:attentions}, the learned attention maps have fairly accurate localization and are able to capture occluded and distant instances (\eg, cars and buildings in the third example). This explains why our model transfers well to detection and segmentation. As the model is trained with open-vocabulary textual annotations, it is able to learn rich visual concepts, some of which (\eg, helmets and goggles) are not even covered in the COCO categories. This shows great potential in the fine-grained localization tasks (such as LVIS~\cite{gupta2019lvis}), which is left as future work. Another interesting direction is to study the zero-shot transfer performance to detection/segmentation based on the learned attention maps.

\myparagraph{Resolution of Mouse Trace Supervision.}

\begin{table}[t]
\setlength{\tabcolsep}{6pt}
\small\centering
\begin{tabular}{lccccc}
    \toprule
    & VOC & \multicolumn{2}{c}{COCO (10\%)} & \multicolumn{2}{c}{COCO (20\%)} \\
    \cmidrule(lr){2-2}\cmidrule(lr){3-4}\cmidrule(lr){5-6}
    Supervision & mAP & $\text{AP}^\text{bbox}$ & $\text{AP}^\text{mask}$ & $\text{AP}^\text{bbox}$ & $\text{AP}^\text{mask}$ \\
    \midrule
    1$\times$ & \textcolor{gray}{87.7} & \textcolor{gray}{24.3} & \textcolor{gray}{21.9} & \textcolor{gray}{29.3} & \textcolor{gray}{25.9} \\
    1$\times$, 2$\times$ & 88.4 & 25.0 & 22.4 & 29.8 & 26.4 \\
    1$\times$, 2$\times$, 4$\times$ & \textcolor{gray}{88.2} & \textcolor{gray}{24.9} & \textcolor{gray}{22.3} & \textcolor{gray}{29.6} & \textcolor{gray}{26.1} \\
    \midrule
    Oracle & \textbf{90.8} & \textbf{26.1} & \textbf{23.3} & \textbf{30.7} & \textbf{21.7} \\
    \bottomrule
\end{tabular}
\caption{Analysis of mouse trace supervision. (1) Applying supervision at too low or too high resolution does not work well. (2) With oracle supervision, the performance is further boosted by \textbf{2\%} on classification and \textbf{1\%} on segmentation.}
\label{tab:analysis:supervisions}
\vspace{-12pt}
\end{table}

We explore different resolutions for mouse trace supervision. As shown in \tab{tab:analysis:supervisions}, applying supervision at both 1$\times$ and 2$\times$ resolutions works the best across different downstream tasks. 1$\times$ alone does not work well due to its low resolution (7$\times$7) while 4$\times$ introduces too much noise from the mouse trace annotation.

\myparagraph{Performance ``Upper Bound''.}

We further investigate the performance upper bound of our method given perfect mouse trace annotations. We synthesize the clean image-caption attention maps using ground-truth COCO segmentation masks. Specifically, we first match each token in the caption with the COCO category names (as well as their synonyms and parent classes). For each token with a match, we compute the intersection-over-union (IoU) between its corresponding mouse trace and every instance mask in the matching category. Finally, we aggregate these instance masks with high IoUs as our oracle image-caption attention maps. The IoU matching process helps to deal with the case where the token in the caption only refers to one of the multiple instances from the category. Note that we apply the oracle supervision still at 1$\times$ and 2$\times$ scale to mimic the coarse resolution of real mouse traces. In \tab{tab:analysis:supervisions}, our \methodshort trained with oracle supervision further pushes the performance by \textbf{2\%} on the PASCAL VOC image classification and \textbf{1\%} on the COCO instance segmentation.

\myparagraph{Training Efficiency.}

In addition to annotation efficiency, our \methodshort pre-training is also very efficient in computation. Its training cost is comparable with ImageNet supervised pre-training. We refer the readers to the appendix for details.
\section{Conclusion}

In this paper, we introduce \methodshort to reduce the practical costs of data annotation by taking advantage of low-cost, multi-modal labels including free-form captions and mouse-over gestures. We adopt a cross-modal contrastive pre-training approach using images and captions, and propose to supervise the image-caption attention map via rendered mouse traces to provide coarse localization information. Extensive experiments verify that the visual features learned through our approach can be effectively and efficiently transferred to downstream tasks including image classification, object detection, and instance segmentation. We hope that our approach will provide a simple but strong baseline and inspire future exploration into how to extract more value from rich yet noisy localized 
textual annotations.

\appendix
\renewcommand{\thesection}{A.\arabic{section}}
\renewcommand{\thefigure}{A\arabic{figure}}
\renewcommand{\thetable}{A\arabic{table}}
\setcounter{section}{0}
\setcounter{figure}{0}
\setcounter{table}{0}

{
\small
\bibliographystyle{ieee}
\bibliography{reference}
}

\clearpage
\onecolumn
\section{Annotation Cost}

\begin{wraptable}{r}{0.55\linewidth}
\vspace{-8pt}
\renewcommand*{\arraystretch}{0.9}
\setlength{\tabcolsep}{5pt}
\small\centering
\begin{tabular}{lccc}
    \toprule
    & Annotations & Cost (hours) & mAP \\
    \midrule
    Multi-label classification & Multi-class labels & 11.1K~\cite{lin2014microsoft,desai2021virtex} & 86.2 \\
    Instance segmentation & Segmentation masks & 30.0K~\cite{lin2014microsoft,desai2021virtex} & 82.3 \\
    \midrule
    LocTex (Ours) & Localized narratives & \textbf{4.7K}~\cite{pont2020connecting} & \textbf{88.4} \\
    \bottomrule
\end{tabular}
\vspace{-8pt}
\caption{Comparison with different forms of annotations.}
\label{tab:annotation}
\vspace{-12pt}
\end{wraptable}

We provide quantitative comparisons between various forms of annotations in \tab{tab:annotation}. Here, all annotation costs are estimated on the 118K training images of COCO. Compared with classification and segmentation annotations, localized narratives are cheaper (lower cost) and offer richer information (higher accuracy). It is worth noticing that the annotation cost of localized narratives is dominated by manual transcription. Thus, its cost can be further reduced by \textbf{3.6$\times$} with an accurate automatic speech recognition system. Annotating over larger sets of classes can be even more challenging since memorizing and learning to distinguish over a large class hierarchy (\eg, 1000 classes for ImageNet) is very costly.

\section{Training Efficiency}

\begin{figure}[h]
\centering
\includegraphics[width=0.9\linewidth]{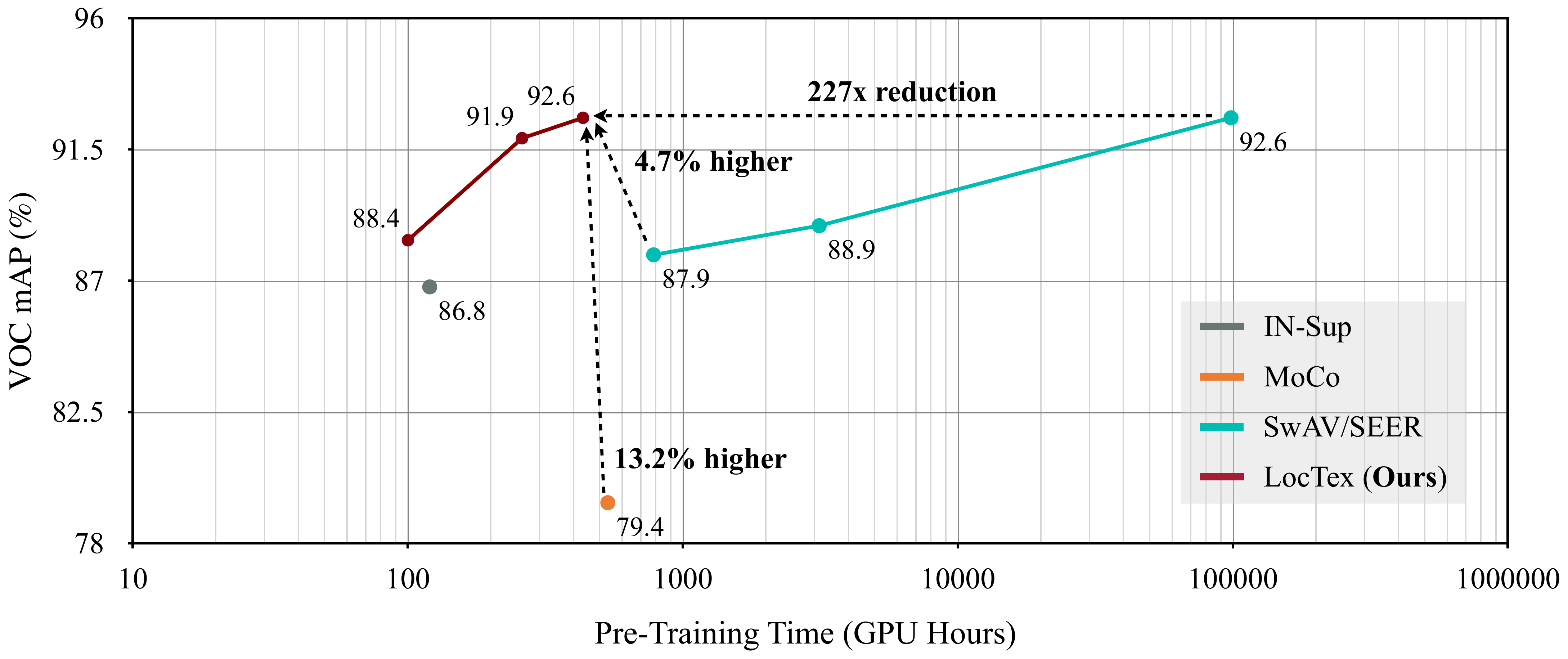}
\end{figure}

Most pre-training efforts focus on improving performance and data efficiency. However, some of them suffer from extremely long training time. We conduct a thorough analysis on the training efficiency across the following pre-training methods:
\begin{itemize}[\hspace{1pt}--]
    \item \textbf{\methodshort}. We include three variants of our \methodshort. One of them is trained only with COCO images for 600 epochs. The remaining two are trained on both COCO and Open Images data, one for 300 epochs and the other one for 500 epochs.
    \item \textbf{SwAV/SEER}~\cite{caron2020unsupervised,goyal2021self}. We include three variants of SwAV as well. Two of them are trained with ImageNet data for 200 and 800 epochs, respectively. The other one is trained with 1 billion uncurated Instagram images (for one epoch).
    \item \textbf{MoCo}~\cite{he2020momentum}. We include the baseline MoCo self-supervised pre-training on ImageNet for 200 epochs.
    \item \textbf{IN-Sup}. We follow the standard ImageNet supervised pre-training (as in \texttt{torchvision}) for 90 epochs.
\end{itemize}
The training time is measured on 8 NVIDIA V100 GPUs for all pre-training methods except SwAV/SEER. For SwAV/SEER, we directly adopt the statistics from their official GitHub repository\footnote{\url{https://github.com/facebookresearch/swav}} and paper~\cite{goyal2021self}: the two trained with ImageNet data use 32 NVIDIA V100 GPUs, and the scaled-up one uses 512 NVIDIA V100 GPUs.

\myparagraph{Results.}

Our \methodshort pre-training is more efficient than ImageNet supervised pre-training while achieving more than 1\% higher accuracy. Compared with the self-supervised pre-training, the improvement is more significant: \ie, we achieve the same linear classification accuracy (92.6) as the scaled-up SwAV with \textbf{227$\times$} less training time. In terms of data efficiency, the scaled-up SwAV requires 1 billion unlabeled images from Instagram while our \methodshort makes use of 809K images with low-cost localized textual annotations. This suggests that supervised pre-training can be much more computationally efficient, and its annotation cost is also affordable if the form of annotation is chosen carefully (which is discussed in the main paper).

\section{Additional Results on COCO Instance Segmentation}

\begin{table*}[t]
\setlength{\tabcolsep}{4pt}
\small\centering
\begin{tabular}{lC{38pt}cccccccccccc}
    \toprule
     & & \multicolumn{6}{c}{50\% Training Data} & \multicolumn{6}{c}{100\% Training Data} \\
    \cmidrule(lr){3-8}\cmidrule(lr){9-14}
     & \# Pretrain Images & $\text{AP}^\text{bbox}$ & $\text{AP}^\text{bbox}_\text{50}$ & $\text{AP}^\text{bbox}_\text{75}$ & $\text{AP}^\text{mask}$ & $\text{AP}^\text{mask}_\text{50}$ & $\text{AP}^\text{mask}_\text{75}$ & $\text{AP}^\text{bbox}$ & $\text{AP}^\text{bbox}_\text{50}$ & $\text{AP}^\text{bbox}_\text{75}$ & $\text{AP}^\text{mask}$ & $\text{AP}^\text{mask}_\text{50}$ & $\text{AP}^\text{mask}_\text{75}$ \\
    \midrule
    Random Init & -- & 28.4 & 46.1 & 29.9 & 25.7 & 43.2 & 26.8 & 36.1 & 55.0 & 38.9 & 31.8 & 52.0 & 33.9 \\
    \midrule
    IN-Sup (10\%) & 128K & 31.2 & 50.0 & 32.7 & 27.9 & 46.8 & 29.2 & 37.7 & 56.9 & 40.6 & 33.0 & 53.4 & 35.3 \\
    VirTex~\cite{desai2021virtex} & 118K & 35.5 & 54.8 & 37.9 & 31.1 & 51.6 & 32.7 & 39.8 & 59.4 & 42.6 & 34.6 & 56.1 & 36.7 \\
    \methodshort (Ours) & 118K & \textbf{36.1} & \textbf{55.7} & \textbf{38.6} & \textbf{31.6} & \textbf{52.4} & \textbf{33.1} & \textbf{40.6} & \textbf{60.6} & \textbf{44.1} & \textbf{35.2} & \textbf{57.0} & \textbf{37.4} \\
    \midrule
    IN-Sup (50\%) & 640K & 34.9 & 54.4 & 37.0 & 30.8 & 51.0 & 32.5 & 39.7 & 59.4 & 43.3 & 34.6 & 56.2 & 36.8 \\
    VirTex~\cite{desai2021virtex} & 118K($\times$5) & 36.6 & 55.9 & 39.3 & 31.9 & 52.6 & 33.6 & 40.8 & 60.5 & 44.2 & 35.2 & 57.0 & 37.6 \\
    \methodshort (Ours) & 809K & \textbf{37.5} & \textbf{57.2} & \textbf{40.4} & \textbf{32.7} & \textbf{54.0} & \textbf{34.7} & \textbf{41.4} & \textbf{61.3} & \textbf{44.9} & \textbf{35.8} & \textbf{57.7} & \textbf{38.4} \\
    \midrule
    IN-Sup (100\%) & 1.28M & 36.3 & 56.1 & 38.8 & 31.9 & 52.7 & 33.7 & 40.2 & 60.0 & 43.5 & 35.0 & 56.4 & 37.4 \\
    \bottomrule
\end{tabular}
\caption{Additional results of instance segmentation on COCO under 50\% and 100\% data settings.}
\label{tab:appendix:coco}
\end{table*}
\begin{figure*}
\centering
\includegraphics[width=0.136\textwidth]{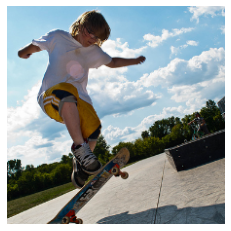}
\includegraphics[width=0.136\textwidth]{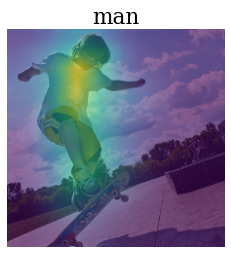}
\includegraphics[width=0.136\textwidth]{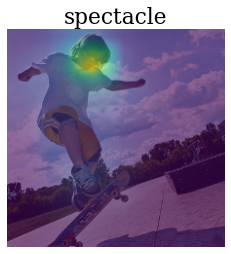}
\includegraphics[width=0.136\textwidth]{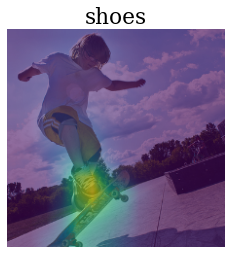}
\includegraphics[width=0.136\textwidth]{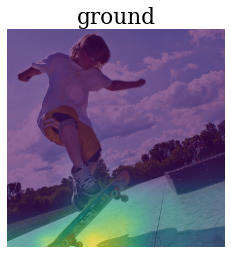}
\includegraphics[width=0.136\textwidth]{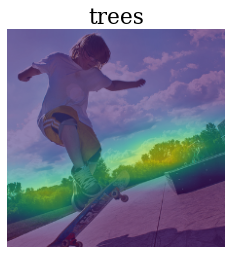}
\includegraphics[width=0.136\textwidth]{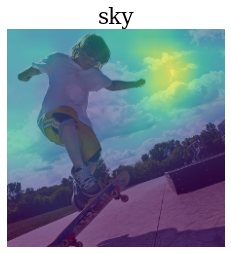}

\includegraphics[width=0.136\textwidth]{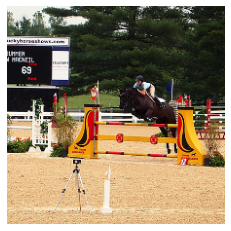}
\includegraphics[width=0.136\textwidth]{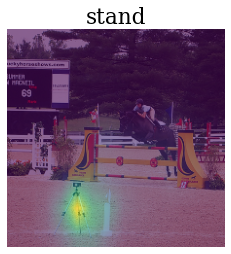}
\includegraphics[width=0.136\textwidth]{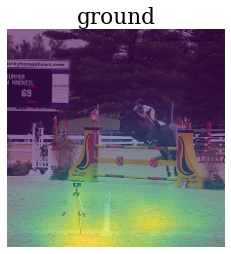}
\includegraphics[width=0.136\textwidth]{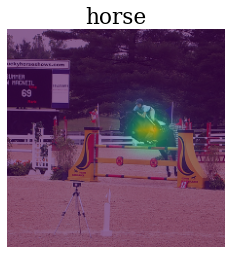}
\includegraphics[width=0.136\textwidth]{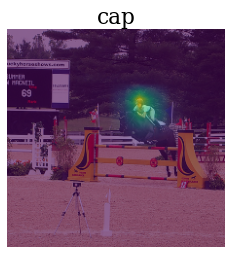}
\includegraphics[width=0.136\textwidth]{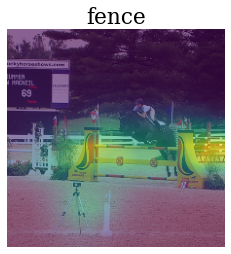}
\includegraphics[width=0.136\textwidth]{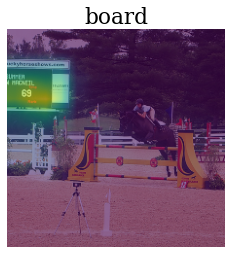}

\includegraphics[width=0.136\textwidth]{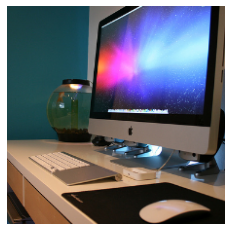}
\includegraphics[width=0.136\textwidth]{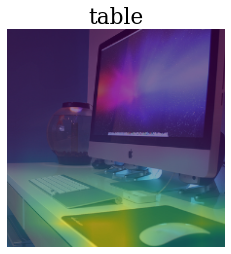}
\includegraphics[width=0.136\textwidth]{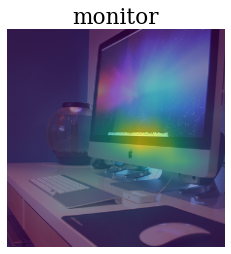}
\includegraphics[width=0.136\textwidth]{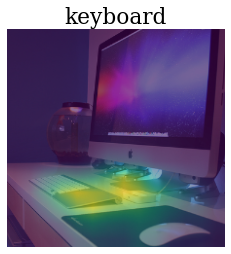}
\includegraphics[width=0.136\textwidth]{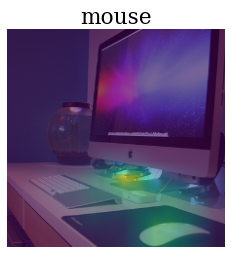}
\includegraphics[width=0.136\textwidth]{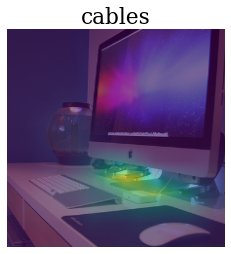}
\includegraphics[width=0.136\textwidth]{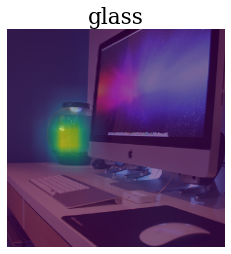}

\includegraphics[width=0.136\textwidth]{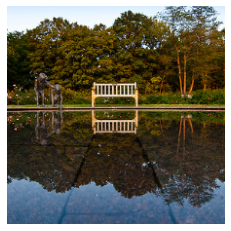}
\includegraphics[width=0.136\textwidth]{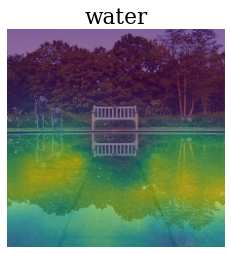}
\includegraphics[width=0.136\textwidth]{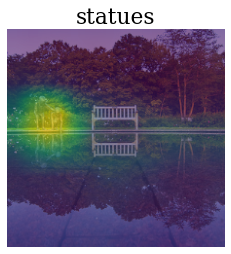}
\includegraphics[width=0.136\textwidth]{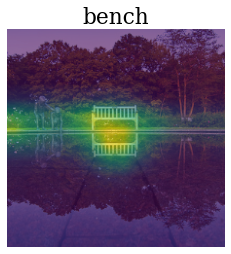}
\includegraphics[width=0.136\textwidth]{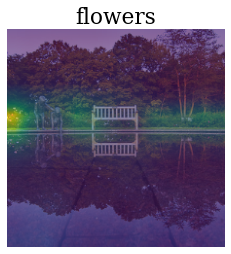}
\includegraphics[width=0.136\textwidth]{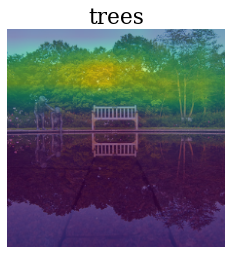}
\includegraphics[width=0.136\textwidth]{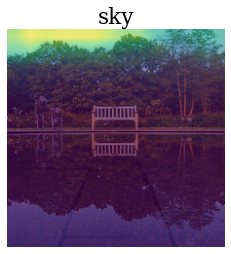}

\caption{Additional visualization of learned image-caption attention maps (on COCO \texttt{val2017}).}
\label{fig:appendix:attentions}
\end{figure*}

In \tab{tab:appendix:coco}, we provide additional results of instance segmentation on COCO under 50\% and 100\% data settings. The experimental setup is exactly the same as in the main paper where we scale the training schedule linearly with the dataset size.

\myparagraph{Results.}

The overall trend is the same as the one under 10\% and 20\% settings (which is presented in the main paper). With the same amount of labelled images, our \methodshort always achieves the highest performance compared with ImageNet supervised pre-training and VirTex pre-training methods. It achieves more than 1\% higher (box or mask) AP than the full ImageNet supervised pre-training baseline while using only half of the annotated images. Under the 100\% data setting, our \methodshort is able to push the instance segmentation performance from 40.2\% to 41.4\% in box AP and from 35.0\% to 35.8\% in mask AP.

\section{Additional Visualizations of Learned Image-Caption Attention Maps}

In \fig{fig:appendix:attentions}, we provide additional visualizations of learned image-caption attention maps on COCO. Note that these visualizations are picked randomly from COCO \texttt{val2017}. The only constraint we apply is to ensure that there are at least six entities in the image for visualization purposes. We observe that the learned attention map is able to localize the instances fairly accurately, even for some small instances (\eg, cap in the second example), which is especially useful for the downstream object detection and instance segmentation tasks.

\end{document}